\documentclass[11pt]{article}
\PassOptionsToPackage{table}{xcolor}
\usepackage[preprint]{acl}
\usepackage{times}
\usepackage{latexsym}
\usepackage[T1]{fontenc}
\usepackage[utf8]{inputenc}
\usepackage{microtype}
\usepackage{amsmath,amssymb}
\usepackage{booktabs,multirow,array}
\usepackage{xcolor}
\usepackage{graphicx}

\definecolor{AVOBlue}{HTML}{123C73}
\definecolor{AVOLight}{HTML}{ECF3FB}
\definecolor{AVOMid}{HTML}{397EC5}
\newcommand{\method}{\mbox{ARP-CD}}
\newcommand{\RMS}{\operatorname{RMS}}
\newcommand{\softmax}{\operatorname{softmax}}

\newcommand{\coauth}{\footnotemark[1]}

\title{Mitigating Video Hallucinations via\\Attention-Guided Representation Perturbation}
\author{
  Zhixin Cai$^{1}$\coauth{},
  Ruoqi Yuan$^{1}$\thanks{Equal contribution.},
  Xinchen Yang$^{1}$,
  Zhuofan Chen$^{1}$,\\
  \textbf{Jun Bai$^{2}$,}
  \textbf{Chuantao Yin$^{3}$,}
  \textbf{Wenge Rong$^{1}$}
  \\[0.5ex]
  \textsuperscript{1}School of Computer Science and Engineering, Beihang University \\
  \textsuperscript{2}Beijing Institute for General Artificial Intelligence, BIGAI \\
  \textsuperscript{3}Sino-French Engineer School, Beihang University
}

\begin{document}
\maketitle

\begin{abstract}
Video large language models (Video LLMs) often generate responses unsupported by the input video.
Contrastive decoding (CD) can mitigate these hallucinations, but its effectiveness depends on how the negative branch weakens visual evidence.
Existing methods rely on predefined perturbations or masking guided by question reconstruction, offering limited control over the model's visual reliance during answer generation.
We introduce Attention-guided Representation Perturbation for Contrastive Decoding (\method{}), a training-free framework for constructing contrastive negatives directly in representation space.
\method{} uses the attention mass assigned to visual tokens at the answer-prediction position to guide bounded gradient updates of the language model's input visual hidden states.
These updates reduce visual attention and produce a negative input, whose predictions are contrasted with those of the original branch throughout generation.
Experiments across four Video LLMs and four benchmarks show that \method{} outperforms representative training-free baselines in most model--benchmark settings, reducing hallucinations while improving general video understanding.
Further analyses show that attention-guided perturbation directions improve negative quality and that the resulting negatives provide effective contrastive references for correcting the original predictions.
\end{abstract}

\section{Introduction}
\label{sec:introduction}

Video LLMs have made substantial progress in video understanding, demonstrating strong performance across tasks such as video question answering, captioning, and reasoning \citep{maaz2024videochatgpt,bai2025qwen}.
Despite this progress, they still exhibit hallucinations, producing responses that are unsupported by or inconsistent with the visual evidence in the input video \citep{wang2024videohallucer,li2025vidhalluc}.
Such hallucinations undermine the reliability of video understanding and can lead to incorrect decisions in downstream applications \citep{zhou2026cctvbench}.
Mitigating hallucinations and strengthening visual grounding are therefore crucial for reliable downstream use, including model-assisted evaluation of generated videos \citep{zhang2024eventhallusion,wu2026omniworldbench}.

Training-based methods mitigate hallucinations through architectural adaptations, contrastive video--text alignment, and reinforcement learning \citep{ma2024vistallama,chang2026santa,zhang2026rewatch}.
However, they incur substantial training costs and additional overhead for curating high-quality supervision \citep{huang2025vistadpo,cai2025taae}.
This motivates training-free methods that bypass these costs through inference-time intervention \citep{huang2024opera,li2025vidhalluc}.
Among these approaches, CD has emerged as an effective strategy for mitigating video hallucinations \citep{wu2025season}.

\begin{figure*}[t]
 \centering
 \includegraphics[width=\textwidth]{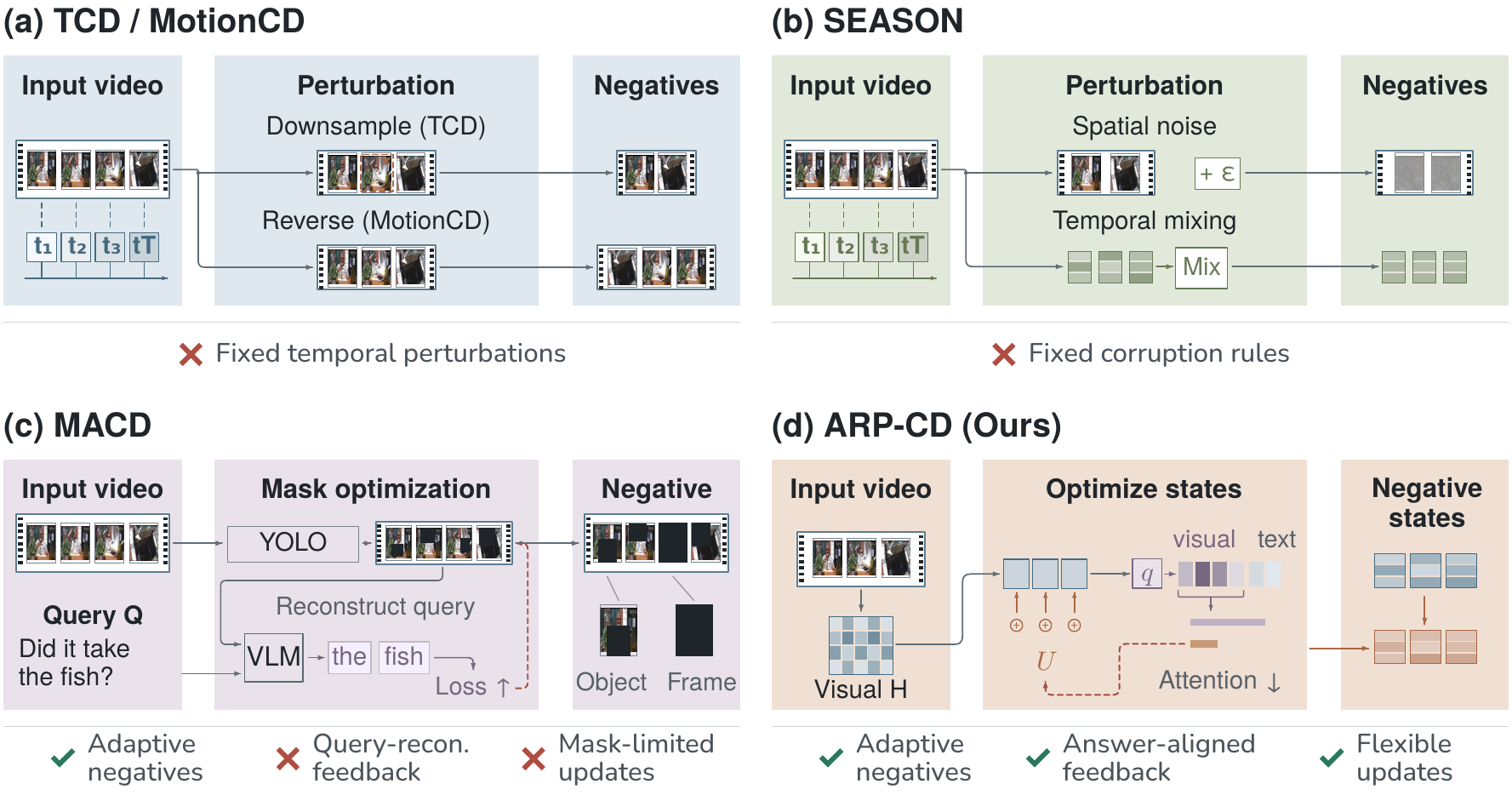}
 \caption{Comparison of negative branch construction in video contrastive decoding.
 (a) TCD downsamples frames, while MotionCD combines predictions from forward and reversed videos \citep{zhang2024eventhallusion,kong2025mhbench}.
 (b) SEASON constructs spatial and temporal negatives through noise injection and temporal feature homogenization \citep{wu2025season}.
 (c) MACD optimizes object and frame masks using query reconstruction feedback \citep{xiao2026macd}.
 (d) \method{} uses attention gradients to optimize visual hidden states, reducing attention to visual tokens at the answer-prediction position.}
 \label{fig:negative_construction}
\end{figure*}

CD favors predictions whose likelihood decreases under weakened visual evidence, making its effectiveness critically dependent on how the negative branch attenuates that evidence \citep{suo2025octopus}.
As shown in Figure~\ref{fig:negative_construction}, existing methods construct negative branches through predefined spatial or temporal perturbations, with limited adaptation to the model's visual dependencies \citep{kong2025mhbench}.
MACD introduces model feedback by using query reconstruction to guide object and frame masking \citep{xiao2026macd}.
However, this reconstruction objective targets question-token prediction rather than directly measuring the model's visual reliance during answer generation \citep{xiao2026macd}.
Moreover, its object and frame masks constrain intervention to content-space transformations, offering less direct control over the hidden representations used for generation than representation-level editing \citep{xiao2026macd,liu2025vti,cai2025taae}.

Visual attention at the generation position provides an internal signal of the model's engagement with visual evidence during answer prediction \citep{huo2025sid,xing2026vader}.
Prior work has associated insufficient visual attention with hallucinated responses, suggesting it captures the visual reliance that CD aims to modulate \citep{huang2024opera,liu2024pai}.
Suppressing attention to informative visual tokens can produce useful contrastive negatives by weakening visual conditioning \citep{huo2025sid,wang2026ascd}.
Task-guided embedding transformations and targeted feature edits further demonstrate that model behavior can be shaped directly in representation space \citep{zhang2025adaqr,cai2026xetrieval}.
In vision--language models, representation-level interventions enable fine-grained adjustments to the features used for generation, including adaptations to model- and input-specific signals \citep{liu2025vti,cai2025taae,wang2026after}.

Building on this insight, we introduce \textbf{\method{}}, a training-free contrastive decoding framework that constructs negative branches through attention-guided representation perturbation.
For each video--prompt pair, \method{} uses the attention mass allocated to visual tokens at the answer-prediction position as an internal optimization signal.
Using the gradients of this signal, it iteratively updates the language model's input visual hidden states within a bounded perturbation budget to reduce visual attention.
The optimized hidden states serve as the input to the negative branch.
CD contrasts predictions from this branch with those of the original branch to favor responses supported by video evidence, thereby mitigating hallucinations.

We evaluate \method{} across four Video LLMs and four benchmarks.
The results show that \method{} outperforms representative training-free baselines in most model--benchmark settings, reducing hallucinations while also improving general video understanding.
Ablation studies show that attention-guided optimization selects perturbation directions that produce more effective contrastive negatives.
Further analyses show that the optimized representations improve generation through the corrective contrast they create with the original branch.
Together, these findings support using the model's visual engagement during answer generation to guide negative construction in representation space.

\section{Related Work}
\label{sec:related_work}

\subsection{Video Hallucination Mitigation}

Video hallucination encompasses unsupported claims about objects and events, incorrect temporal relations, and facts beyond the video evidence, motivating dedicated evaluation in VideoHallucer, EventHallusion, and VidHalluc \citep{wang2024videohallucer,zhang2024eventhallusion,li2025vidhalluc}.
Training-based approaches improve grounding through changes to visual attention and temporal aggregation, or through contrastive video--text alignment \citep{ma2024vistallama,chang2026santa}.
VistaDPO and PaMi-VDPO use hierarchical and prompt-aware preference learning, respectively, to improve video--response alignment \citep{huang2025vistadpo,ding2025pamivdpo}.
Training-free approaches intervene during inference; DINO-HEAL, for example, reweights visual features using spatial saliency from DINOv2 \citep{li2025vidhalluc}.
Our work studies negative construction for contrastive decoding with frozen Video LLMs.

\subsection{Contrastive Decoding for Video LLMs}

Contrastive decoding uses differences between model predictions to adjust token preferences, with VCD applying this principle to original and noise-corrupted images \citep{li2023contrastive,leng2024vcd}.
Video methods adapt negative construction to temporal information, with TCD downsampling frames and MotionCD combining predictions from forward and reversed videos to suppress motion cues \citep{zhang2024eventhallusion,kong2025mhbench}.
SEASON constructs temporal negatives through feature homogenization and spatial negatives through noise injection, then uses attention divergence to weight their contrastive contributions for each generated token \citep{wu2025season}.
MACD optimizes object and frame masking strengths using gradients of query reconstruction loss, adapting the counterfactual video to the model and question \citep{xiao2026macd}.
\method{} optimizes the negative in visual representation space, using attention at the answer-prediction position as the signal for bounded updates of the input visual hidden states.

\subsection{Attention and Representation Intervention}

In image-based LVLMs, PAI increases attention to image tokens, while SID retains less informative visual tokens and ASCD directly steers attention to form contrastive predictions \citep{liu2024pai,huo2025sid,wang2026ascd}.
For Video LLMs, VADER adaptively reallocates attention and constructs a contrastive branch by erasing important visual tokens within each frame \citep{xing2026vader}.

Representation editing offers another form of inference-time intervention.
VTI steers visual and textual latent representations, while AFTER combines factual steering vectors with query-adaptive offsets \citep{liu2025vti,wang2026after}.
In video models, TA-AE selects and edits hallucination-sensitive internal modules according to temporal variation \citep{cai2025taae}.
\method{} uses attention as a differentiable objective for per-input visual-state perturbations, integrating attention guidance, bounded representation updates, and contrastive decoding in a single procedure.

\section{Method}
\label{sec:method}

\begin{figure*}[t]
 \centering
 \includegraphics[width=\textwidth]{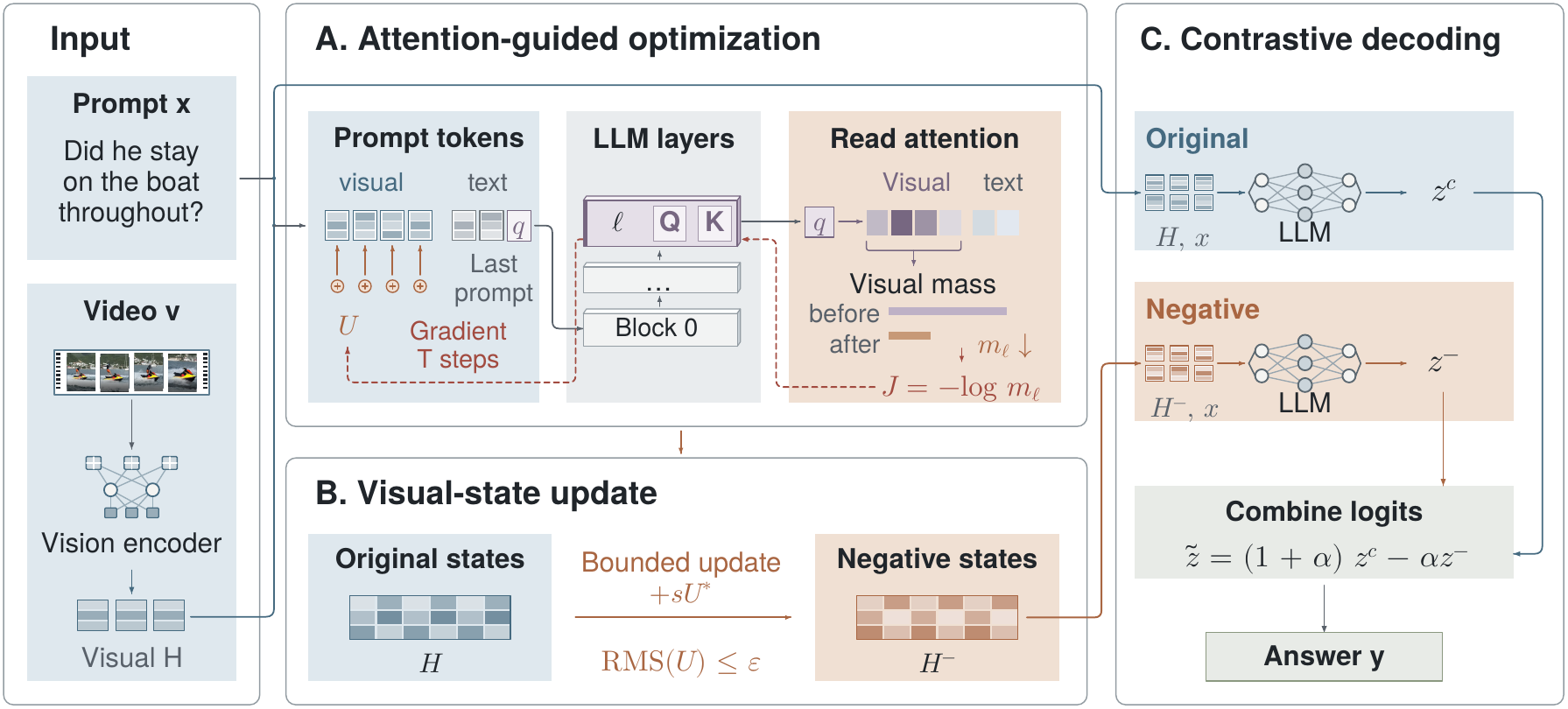}
 \caption{Overview of \method{}.
 (A) We reduce attention from the last prompt token to visual patches by backpropagating the attention objective from a selected decoder layer to the input perturbation.
 (B) Bounded perturbations of the input visual hidden states produce a negative representation.
 (C) CD combines predictions from the original and negative branches.}
 \label{fig:method_overview}
\end{figure*}

\method{} combines attention-guided representation perturbation (ARP) with contrastive decoding (CD), as shown in Figure~\ref{fig:method_overview}.
For each video--prompt pair, ARP constructs a negative representation by optimizing the input visual hidden states to reduce visual attention at the answer-prediction position.
CD then combines predictions from the original and optimized representations throughout answer generation.

\subsection{Preliminaries}
\label{sec:preliminaries}

\paragraph{Video-conditioned generation.}
Let $v$ be a video and $x$ its text prompt.
The visual encoder and projection or merger module map the frames to visual hidden states $H\in\mathbb{R}^{N\times d}$ at the language-model input, where $N$ is the number of visual patch tokens and $d$ is the hidden dimension.
The model with parameters $\theta$ generates an answer $y$ autoregressively:
\begin{equation}
 p_\theta(y\mid H,x)=\prod_{t=1}^{|y|}p_\theta(y_t\mid H,x,y_{<t}).
\end{equation}

\paragraph{Contrastive decoding.}
Contrastive decoding adjusts token preferences using predictions from an original branch and an auxiliary negative branch.
Let $H^-$ denote the negative visual representation, and let $z_t^c$ and $z_t^-$ be the logits obtained from $H$ and $H^-$ under the same prompt $x$ and generated prefix $y_{<t}$.
The contrastive score is
\begin{equation}
 \widetilde z_t(w)=(1+\alpha)z_t^c(w)-\alpha z_t^-(w),
 \label{eq:cd}
\end{equation}
where $\alpha\geq0$ controls the correction strength.
A plausibility constraint retains tokens supported by the original distribution $p_t^c=\softmax(z_t^c)$:
\begin{equation}
 \mathcal C_t=\left\{w:\frac{p_t^c(w)}{\max_{u}p_t^c(u)}\geq\beta\right\},\,0<\beta\leq1.
 \label{eq:mask}
\end{equation}

\subsection{Visual Attention Objective}
\label{sec:objective}

We use attention to visual patches as an internal proxy for visual engagement during answer prediction.
For a candidate representation $H'$, we process the complete multimodal prompt and observe attention at decoder layer $\ell$ for the last valid prompt position $q$, which is used to predict the first answer token.
Let $\mathcal K_q$ be the key positions allowed by the model's native attention mask, and $\mathcal V\subseteq\mathcal K_q$ the visible visual patch positions.
For head $h$, with query $Q_q^{(\ell,h)}$, key $K_j^{(\ell,h)}$, and head dimension $d_h$, the attention weights are
\begin{equation}
 A_{qj}^{(\ell,h)}=\softmax_{j\in\mathcal K_q}
 \left(\frac{Q_q^{(\ell,h)}K_j^{(\ell,h)\top}}{\sqrt{d_h}}\right).
 \label{eq:attention}
\end{equation}
We define $m_\ell$ as the fraction of attention allocated to visual patches, averaged over all $n_h$ query heads:
\begin{equation}
 m_\ell(H',x)=\frac{1}{n_h}\sum_{h=1}^{n_h}\sum_{j\in\mathcal V}A_{qj}^{(\ell,h)}.
 \label{eq:mass}
\end{equation}

\method{} seeks a nearby representation with lower visual attention by maximizing
\begin{equation}
 J(H';x)=-\log m_\ell(H',x).
 \label{eq:objective}
\end{equation}
Maximizing $J$ reduces the mean visual attention mass; the logarithm is applied after averaging across heads.

\subsection{Attention-Guided Representation Perturbation}
\label{sec:optimization}

\paragraph{Parameterization.}
We parameterize the candidate visual states as
\begin{equation}
 s=\RMS(H),\qquad H(U)=H+sU,
 \label{eq:parameterization}
\end{equation}
where $\RMS(X)=\sqrt{\frac{1}{|X|}\sum_iX_i^2}$ and $|X|$ is the number of scalar entries in $X$.
We optimize the dimensionless perturbation $U\in\mathbb R^{N\times d}$ subject to a budget $\epsilon>0$:
\begin{equation}
 \max_{U:\,\RMS(U)\leq\epsilon}J(H+sU;x).
 \label{eq:problem}
\end{equation}
We impose this relative RMS constraint to limit excessive representational distortion and maintain an informative contrast with the original branch.
Text and special-token embeddings are excluded from the perturbation.

\begin{table*}[t]
\centering
\small
\setlength{\tabcolsep}{7pt}
\renewcommand{\arraystretch}{1.1}
\begin{tabular}{clcccc}
\toprule
\multirow{2}{*}{Model} & \multirow{2}{*}{Method} & \multicolumn{2}{c}{Hallucination} & \multicolumn{2}{c}{General Understanding} \\
\cmidrule(lr){3-4}\cmidrule(lr){5-6}
 & & VideoHallucer & EventHallusion & Video-MME & MVBench \\
\midrule
\multirow{6}{*}{\textbf{Qwen2.5-VL-3B}} & Vanilla & 46.51 & 40.23 & 51.85 & 58.83 \\
 & DINO-HEAL & \cellcolor{AVOMid!10}\underline{47.31} & 40.65 & 51.89 & 59.18 \\
 & TCD & 43.52 & 43.04 & \cellcolor{AVOMid!10}\underline{52.00} & \cellcolor{AVOMid!10}\underline{59.40} \\
 & MotionCD & 36.35 & \cellcolor{AVOMid!10}\underline{43.46} & 45.44 & 52.95 \\
 & MACD & 44.31 & 40.93 & 50.56 & 57.63 \\
 & \textbf{\method{} (Ours)} & \cellcolor{AVOMid!30} \textbf{48.01} & \cellcolor{AVOMid!30} \textbf{51.48} & \cellcolor{AVOMid!30} \textbf{52.48} & \cellcolor{AVOMid!30} \textbf{59.55} \\
\midrule
\multirow{6}{*}{\textbf{Qwen2.5-VL-7B}} & Vanilla & 51.96 & 49.23 & 53.52 & 60.40 \\
 & DINO-HEAL & \cellcolor{AVOMid!10}\underline{52.26} & 49.51 & 53.89 & 60.50 \\
 & TCD & 51.62 & 51.34 & \cellcolor{AVOMid!10}\underline{53.93} & \cellcolor{AVOMid!10}\underline{60.60} \\
 & MotionCD & 44.39 & \cellcolor{AVOMid!30} \textbf{52.74} & 46.48 & 54.35 \\
 & MACD & 47.33 & 51.20 & 52.35 & 58.68 \\
 & \textbf{\method{} (Ours)} & \cellcolor{AVOMid!30} \textbf{53.30} & \cellcolor{AVOMid!10}\underline{52.32} & \cellcolor{AVOMid!30} \textbf{54.59} & \cellcolor{AVOMid!30} \textbf{60.73} \\
\midrule
\multirow{6}{*}{\textbf{Video-LLaVA-7B}} & Vanilla & 27.93 & 23.35 & 26.19 & \cellcolor{AVOMid!10}\underline{43.68} \\
 & DINO-HEAL & \cellcolor{AVOMid!10}\underline{28.23} & 23.35 & 27.30 & 43.30 \\
 & TCD & 27.00 & 24.75 & 26.56 & 42.63 \\
 & MotionCD & 24.09 & \cellcolor{AVOMid!30} \textbf{36.57} & \cellcolor{AVOMid!10}\underline{28.94} & 40.45 \\
 & MACD & 27.73 & 25.88 & 28.64 & 42.53 \\
 & \textbf{\method{} (Ours)} & \cellcolor{AVOMid!30} \textbf{31.95} & \cellcolor{AVOMid!10}\underline{35.16} & \cellcolor{AVOMid!30} \textbf{32.91} & \cellcolor{AVOMid!30} \textbf{43.88} \\
\midrule
\multirow{6}{*}{\textbf{LLaVA-NeXT-Video-7B}} & Vanilla & \cellcolor{AVOMid!30} \textbf{26.41} & 32.91 & 29.89 & 43.55 \\
 & DINO-HEAL & \cellcolor{AVOMid!10}\underline{26.21} & 33.19 & 30.00 & 42.58 \\
 & TCD & 24.51 & 34.88 & 30.96 & 43.73 \\
 & MotionCD & 22.60 & 34.46 & \cellcolor{AVOMid!10}\underline{32.16} & 41.13 \\
 & MACD & 23.21 & \cellcolor{AVOMid!10}\underline{38.54} & 31.65 & \cellcolor{AVOMid!30} \textbf{44.80} \\
 & \textbf{\method{} (Ours)} & 25.72 & \cellcolor{AVOMid!30} \textbf{52.18} & \cellcolor{AVOMid!30} \textbf{37.04} & \cellcolor{AVOMid!10}\underline{44.48} \\
\bottomrule
\end{tabular}
\caption{Main results (\%).
EventHallusion reports the count-weighted aggregate of selection and description scores.
Within each model and benchmark, the best scores are bold with darker blue shading, and the second-best are underlined with lighter blue shading.}
\label{tab:main}
\end{table*}

\paragraph{Projected gradient updates.}
We backpropagate the attention objective from layer $\ell$ to the input perturbation $U$.
Starting from $U_0=0$, we perform up to $T$ steps of RMS-normalized projected gradient ascent:
\begin{align}
 G_k&=\nabla_{U_k}J(H+sU_k;x),\label{eq:gradient}\\
 \widehat U_{k+1}&=U_k+\eta\frac{G_k}{\RMS(G_k)},\label{eq:update}\\
 U_{k+1}&=\frac{\widehat U_{k+1}}{\max\left(1,\RMS(\widehat U_{k+1})/\epsilon\right)},
 \label{eq:projection}
\end{align}
where $\eta$ is the step size.
RMS normalization makes the update scale independent of the raw gradient magnitude, while projection keeps the accumulated perturbation within the budget.
We form the negative representation $H^-=H+sU^\star$ using the iterate with the highest objective value, retaining vanilla decoding if optimization yields no improvement.

\subsection{Contrastive Generation}
\label{sec:generation}

For each video--prompt pair, we optimize the negative representation $H^-$ once and keep the input perturbation fixed throughout generation.
At each decoding step, we combine the logits of the original and negative branches using Eq.~\ref{eq:cd} and restrict candidates using Eq.~\ref{eq:mask}.
After applying the checkpoint's decoding processors, we select the highest-scoring token and append it to the shared answer prefix.

\section{Experiments}
\label{sec:experiments}

\subsection{Experimental Setup}
\label{sec:setup}

\paragraph{Backbones.}
We evaluate four Video LLMs spanning different model scales and architectures: Qwen2.5-VL-3B and -7B \citep{bai2025qwen}, Video-LLaVA-7B \citep{lin2024videollava}, and LLaVA-NeXT-Video-7B \citep{zhang2024llavanext}.

\paragraph{Benchmarks.}
We evaluate hallucination mitigation on VideoHallucer (VH; \citealt{wang2024videohallucer}), which covers five hallucination categories, and EventHallusion (EH; \citealt{zhang2024eventhallusion}), which assesses event hallucinations through selection and description tasks.
For general video understanding, Video-MME \citep{fu2025videomme} spans short, medium, and long videos, while MVBench \citep{li2024mvbench} covers 20 diverse tasks.

\paragraph{Baselines.}
We compare \method{} with vanilla decoding and four representative training-free baselines: DINO-HEAL \citep{li2025vidhalluc}, TCD \citep{zhang2024eventhallusion}, MotionCD \citep{kong2025mhbench}, and MACD \citep{xiao2026macd}.
These cover saliency-based feature enhancement, temporal perturbations, and mask-based negative construction.

\begin{table*}[t]
\centering
\small
\setlength{\tabcolsep}{7pt}
\renewcommand{\arraystretch}{1.1}
\begin{tabular}{clcccc}
\toprule
Model & Method & VideoHallucer & EventHallusion & Video-MME & MVBench \\
\midrule
\multirow{5}{*}{\textbf{Qwen2.5-VL-3B}} & IsoPert & 46.10 & 41.35 & 51.89 & 59.00 \\
 & RadCon & 46.60 & 41.35 & 51.96 & 58.85 \\
 & QueryRec & 46.25 & 38.68 & 52.15 & 59.23 \\
 & AttnCal & 46.92 & 38.12 & 51.74 & 58.93 \\
 & \textbf{\method{} (Ours)} & \cellcolor{AVOMid!30} \textbf{48.01} & \cellcolor{AVOMid!30} \textbf{51.48} & \cellcolor{AVOMid!30} \textbf{52.48} & \cellcolor{AVOMid!30} \textbf{59.55} \\
\midrule
\multirow{5}{*}{\textbf{Video-LLaVA-7B}} & IsoPert & 29.23 & 23.63 & 27.22 & 43.68 \\
 & RadCon & 29.15 & 22.36 & 27.11 & 43.65 \\
 & QueryRec & 29.01 & 26.72 & 29.34 & 43.65 \\
 & AttnCal & 27.11 & 26.02 & 27.33 & 43.65 \\
 & \textbf{\method{} (Ours)} & \cellcolor{AVOMid!30} \textbf{31.95} & \cellcolor{AVOMid!30} \textbf{35.16} & \cellcolor{AVOMid!30} \textbf{32.91} & \cellcolor{AVOMid!30} \textbf{43.88} \\
\bottomrule
\end{tabular}
\caption{Ablation results for negative construction (\%).
Within each model and benchmark, the best scores are bold with blue shading.}
\label{tab:ablation}
\end{table*}

\subsection{Main Results}
\label{sec:main_results}

Table~\ref{tab:main} compares \method{} with the baselines across four models and four benchmarks.
\method{} achieves the best score in 12 of the 16 model--benchmark settings and ranks second in three others, demonstrating its effectiveness across models and tasks.

\paragraph{Hallucination mitigation.}
On VideoHallucer, \method{} exceeds the strongest baseline by 0.70, 1.04, and 3.71 percentage points (pp) for Qwen2.5-VL-3B, Qwen2.5-VL-7B, and Video-LLaVA, respectively.
For LLaVA-NeXT, all evaluated interventions underperform vanilla decoding on VideoHallucer.
On EventHallusion, it leads for Qwen2.5-VL-3B and LLaVA-NeXT, with gains of 8.02 and 13.64 pp over the strongest baseline, and ranks second for the other two models.

\paragraph{General video understanding.}
\method{} achieves the highest Video-MME score on every backbone, improving over the strongest baseline by 0.48--4.88 pp.
On MVBench, it leads for both Qwen sizes and Video-LLaVA and ranks second for LLaVA-NeXT, with gains of 0.20--0.93 pp over Vanilla.

Together, these results demonstrate the effectiveness of attention-guided negative construction across model architectures, supporting both hallucination mitigation and general video understanding.

\subsection{Attention-Guided Negative Construction}
\label{sec:ablation}

To understand what makes \method{}'s negatives effective, we isolate the roles of perturbation direction, optimization objective, and attention reduction.

\paragraph{Perturbation direction.}
\textit{Isotropic Perturbation (IsoPert)} replaces the optimized direction with a Gaussian direction, while \textit{Radial Contraction (RadCon)} scales the visual states toward the origin.
Both match \method{}'s realized relative RMS separately for each example, isolating the effect of direction at a fixed displacement magnitude.

\paragraph{Optimization objective.}
\textit{Query Reconstruction (QueryRec)} replaces the attention objective with MACD's query-reconstruction loss, maximizing the mean next-token negative log-likelihood of the input instruction \citep{xiao2026macd}.
This comparison tests whether targeting visual reliance during answer generation yields more effective negatives than optimizing question reconstruction.

\begin{table*}[t]
\centering
\small
\setlength{\tabcolsep}{6.5pt}
\renewcommand{\arraystretch}{1.1}
\begin{tabular}{llcccc}
\toprule
\multirow{2}{*}{Attention modulation} & \multirow{2}{*}{Generation} & \multicolumn{2}{c}{\textbf{Qwen2.5-VL-3B}} & \multicolumn{2}{c}{\textbf{Video-LLaVA-7B}} \\
\cmidrule(lr){3-4}\cmidrule(lr){5-6}
 & & VideoHallucer & Video-MME & VideoHallucer & Video-MME \\
\midrule
Unmodified & Direct & 46.51 & 51.85 & 27.93 & 26.19 \\
Suppress ($\downarrow$) & \textbf{Contrast (\method{}, Ours)} & \cellcolor{AVOMid!30} \textbf{48.01} & \cellcolor{AVOMid!30} \textbf{52.48} & \cellcolor{AVOMid!30} \textbf{31.95} & \cellcolor{AVOMid!30} \textbf{32.91} \\
Enhance ($\uparrow$) & Contrast & 44.17 & 51.96 & 29.04 & 26.46 \\
Suppress ($\downarrow$) & Direct & 45.02 & 51.21 & 27.86 & 26.08 \\
Enhance ($\uparrow$) & Direct & 45.73 & 51.33 & 25.80 & 28.89 \\
\bottomrule
\end{tabular}
\caption{Attention modulation and generation strategy (\%).
Contrast uses the optimized representations as a negative branch; Direct generates from them alone.
Within each model and benchmark, the best scores are bold with blue shading.}
\label{tab:direction}
\end{table*}

\paragraph{Attention reduction.}
\textit{Attention Calibration (AttnCal)} subtracts a scalar bias from visual-key attention logits at layer $\ell$.
The bias is calibrated to match \method{}'s visual attention mass at position $q$ for each example and remains fixed during negative-branch decoding.
This separates the observed attention reduction from the representation updates used to achieve it.

Table~\ref{tab:ablation} shows that \method{} outperforms all four controls.
The comparisons with IsoPert and RadCon demonstrate the value of attention-guided directions.
The advantage over QueryRec further supports using answer-position attention to guide representation optimization.
AttnCal matches the attention target but yields lower scores, showing that the representation changes induced by optimization contribute beyond the resulting attention mass.

Together, these comparisons show that attention-guided directions improve negative quality beyond perturbation magnitude and attention reduction.

\begin{table*}[t]
\centering
\small
\setlength{\tabcolsep}{7pt}
\renewcommand{\arraystretch}{1.1}
\begin{tabular}{clcccc}
\toprule
\multirow{2}{*}{Model} & \multirow{2}{*}{Method} & \multicolumn{2}{c}{VideoHallucer} & \multicolumn{2}{c}{Video-MME} \\
\cmidrule(lr){3-4}\cmidrule(lr){5-6}
 & & Retained (\%) & Mean $D$ & Retained (\%) & Mean $D$ \\
\midrule
\multirow{5}{*}{\textbf{Qwen2.5-VL-3B}} & IsoPert & 100.00 & -0.0025 & 100.01 & -0.0030 \\
 & RadCon & 100.28 & -0.0031 & 100.15 & -0.0014 \\
 & QueryRec & 103.09 & -0.0019 & 126.02 & 0.0102 \\
 & AttnCal & 44.42 & -0.0036 & 9.69 & -0.0038 \\
 & \textbf{\method{} (Ours)} & 44.42 & \cellcolor{AVOMid!30} \textbf{0.0175} & 9.69 & \cellcolor{AVOMid!30} \textbf{0.0208} \\
\midrule
\multirow{5}{*}{\textbf{Video-LLaVA-7B}} & IsoPert & 100.00 & -0.0002 & 100.00 & 0.0001 \\
 & RadCon & 100.18 & -0.0002 & 100.20 & 0.0003 \\
 & QueryRec & 97.34 & -0.0055 & 101.81 & 0.0244 \\
 & AttnCal & 59.12 & -0.0019 & 54.75 & 0.0004 \\
 & \textbf{\method{} (Ours)} & 59.12 & \cellcolor{AVOMid!30} \textbf{0.0046} & 54.75 & \cellcolor{AVOMid!30} \textbf{0.0722} \\
\bottomrule
\end{tabular}
\caption{First-token contrastive effects on candidate pairs retained by the original branch's plausibility mask.
Retained is the median negative-to-original visual attention ratio; mean $D$ measures the logit-margin correction per unit $\alpha$ in raw logit units.
The largest mean $D$ values are bold with blue shading.}
\label{tab:contrastive_effects}
\end{table*}

\subsection{Attention Modulation and Generation Strategy}
\label{sec:factorial}

We next separate the effect of changing visual attention from the way the optimized representation is used for generation.
We cross attention suppression or enhancement with contrastive decoding or direct generation.
Unlike \method{}, which maximizes $-\log m_\ell$ to suppress visual attention, the enhancement condition maximizes $+\log m_\ell$ to increase it.
Direct generation decodes from the optimized states alone.
Vanilla decoding serves as the common reference.
Table~\ref{tab:direction} summarizes the results.

In all settings, directly generating from attention-suppressed representations underperforms Vanilla, whereas using them as the negative branch in CD (\method{}) outperforms all other conditions.
This reversal shows that attention-guided degradation creates an informative contrast that CD can exploit to correct the original predictions.

Although using the attention-enhanced branch for CD improves over Vanilla in three of the four settings, direct generation from the same representations lowers performance in most settings.
This seemingly counterintuitive result suggests that increasing total visual attention does not necessarily improve the use of relevant evidence.
A plausible explanation is that the additional attention is allocated to visual content unrelated to the question, interfering with answer generation \citep{kang2025visualsink}.
The enhanced branch can still provide a useful contrastive reference for correcting the original predictions.

These results support \method{}'s use of attention-suppressed representations as corrective references, showing that their contrastive value differs from their standalone predictive quality.

\begin{table*}[t]
\centering
\small
\setlength{\tabcolsep}{7pt}
\renewcommand{\arraystretch}{1.1}
\begin{tabular}{clcccc}
\toprule
\multirow{2}{*}{Model} & \multirow{2}{*}{Method} & \multicolumn{2}{c}{VideoHallucer} & \multicolumn{2}{c}{Video-MME} \\
\cmidrule(lr){3-4}\cmidrule(lr){5-6}
 & & Time (s) & $\times$Vanilla & Time (s) & $\times$Vanilla \\
\midrule
\multirow{6}{*}{\textbf{Qwen2.5-VL-3B}} & Vanilla & 0.46 & 1.00 & 0.84 & 1.00 \\
 & DINO-HEAL & 0.68 & 1.39 & 1.22 & 1.44 \\
 & TCD & 1.09 & 2.27 & 1.74 & 2.09 \\
 & MotionCD & 1.33 & 3.12 & 2.58 & 3.00 \\
 & MACD & 8.84 & 22.66 & 19.66 & 19.84 \\
\rowcolor{AVOLight}  & \textbf{\method{} (Ours)} & 1.58 & 3.76 & 4.27 & 4.87 \\
\midrule
\multirow{6}{*}{\textbf{Video-LLaVA-7B}} & Vanilla & 0.34 & 1.00 & 0.50 & 1.00 \\
 & DINO-HEAL & 0.39 & 1.10 & 0.52 & 1.06 \\
 & TCD & 0.70 & 2.02 & 1.00 & 2.26 \\
 & MotionCD & 1.06 & 2.91 & 1.42 & 2.90 \\
 & MACD & 8.57 & 20.84 & 9.24 & 17.68 \\
\rowcolor{AVOLight}  & \textbf{\method{} (Ours)} & 2.42 & 6.25 & 2.60 & 5.33 \\
\bottomrule
\end{tabular}
\caption{Inference latency on sampled inputs using a single A100 40GB GPU.
Time is the median latency per answer; $\times$Vanilla is the median per-input latency ratio to Vanilla.}
\label{tab:cost}
\end{table*}

\subsection{Mechanistic Analysis}
\label{sec:mechanism}

We examine how representation perturbation shapes the logit correction in contrastive decoding, comparing \method{} with the ablation controls.
For each question, let $a$ be the correct answer token and $b$ the highest-scoring incorrect answer token under the original branch.
We use the same pair across methods and define its \emph{logit margin} as the score of $a$ minus that of $b$.
For the original and negative branches, respectively,
\begin{equation}
 \begin{aligned}
 M &= z_1^c(a)-z_1^c(b),\\
 M^- &= z_1^-(a)-z_1^-(b).
 \end{aligned}
 \label{eq:branch_margins}
\end{equation}
A positive margin indicates that the correct candidate has the higher score.
We measure the contribution of the negative branch using $D=M-M^-$.
By Eq.~\ref{eq:cd}, the margin after CD is $M_{\mathrm{CD}}=M+\alpha D$, so positive $D$ increases the correct candidate's relative score.
We analyze examples where both selected tokens pass the original branch's plausibility mask, focusing on margin corrections between candidates that remain eligible for decoding.

Table~\ref{tab:contrastive_effects} reports mean $D$ and median attention retention.
Across all settings, \method{} yields positive mean $D$ values exceeding those of all ablation controls, indicating a stronger average contrastive correction toward the correct answer.

\subsection{Computational Cost}
\label{sec:cost}

To quantify the inference overhead of \method{}, we compare inference latency with the baselines on a single NVIDIA A100 40GB GPU.
All methods use the same inputs sampled from VideoHallucer and Video-MME.

Table~\ref{tab:cost} shows that \method{} runs faster than MACD in every setting but slower than the other baselines.
The additional latency over non-iterative baselines comes from the forward and backward passes used to optimize each negative.
By reusing encoded visual representations, \method{} avoids the object detection and repeated visual encoding involved in MACD's mask optimization.
These results show that \method{} enables more effective model-guided negative construction at lower inference latency.

\section{Conclusion}
\label{sec:conclusion}

We presented \method{}, a training-free framework for mitigating video hallucinations through attention-guided representation perturbation.
Experiments demonstrate improvements over representative training-free baselines in most settings, spanning hallucination mitigation and general video understanding.
Controlled studies show that negative quality depends on the direction of representation perturbation and that the optimized negatives improve generation through their contrast with the original predictions.
These findings demonstrate that attention-guided representation perturbation provides effective contrastive signals for mitigating video hallucinations.

\section*{Limitations}

Although \method{} achieves substantially lower latency than MACD for model-guided negative construction, its iterative forward and backward passes still incur higher latency than non-iterative baselines, motivating more efficient per-input optimization.
The method also requires access to internal attention computations and gradients, limiting its use with black-box model services.

\method{} remains dependent on the pretrained backbone's visual perception capabilities.
Errors caused by missed or poorly encoded visual evidence may therefore persist after contrastive decoding.

\section*{Ethical Considerations}

\method{} aims to improve the reliability of video understanding by reducing hallucinations in pretrained Video LLMs.
Our experiments use publicly available datasets, pretrained models, and third-party software in accordance with their respective licenses and terms of use.

Generative AI tools assisted with code development and analysis, literature search, manuscript organization, drafting and editing, and figure preparation.
The authors retain full responsibility for the methods, results, citations, and final manuscript.

\bibliography{references}

@inproceedings{li2023contrastive,
  title={Contrastive decoding: Open-ended text generation as optimization},
  author={Li, Xiang Lisa and Holtzman, Ari and Fried, Daniel and Liang, Percy and Eisner, Jason and Hashimoto, Tatsunori B and Zettlemoyer, Luke and Lewis, Mike},
  booktitle={Proceedings of the 61st annual meeting of the association for computational linguistics (volume 1: Long papers)},
  pages={12286--12312},
  year={2023}
}

@inproceedings{leng2024vcd,
  title={Mitigating object hallucinations in large vision-language models through visual contrastive decoding},
  author={Leng, Sicong and Zhang, Hang and Chen, Guanzheng and Li, Xin and Lu, Shijian and Miao, Chunyan and Bing, Lidong},
  booktitle={2024 IEEE/CVF Conference on Computer Vision and Pattern Recognition (CVPR)},
  pages={13872--13882},
  year={2024},
  organization={IEEE}
}

@misc{bai2025qwen,
      title={Qwen2.5-VL Technical Report}, 
      author={Shuai Bai and Keqin Chen and Xuejing Liu and Jialin Wang and Wenbin Ge and Sibo Song and Kai Dang and Peng Wang and Shijie Wang and Jun Tang and Humen Zhong and Yuanzhi Zhu and Mingkun Yang and Zhaohai Li and Jianqiang Wan and Pengfei Wang and Wei Ding and Zheren Fu and Yiheng Xu and Jiabo Ye and Xi Zhang and Tianbao Xie and Zesen Cheng and Hang Zhang and Zhibo Yang and Haiyang Xu and Junyang Lin},
      year={2025},
      eprint={2502.13923},
      archivePrefix={arXiv},
      primaryClass={cs.CV},
      url={https://arxiv.org/abs/2502.13923}, 
}

@inproceedings{lin2024videollava,
  title={Video-llava: Learning united visual representation by alignment before projection},
  author={Lin, Bin and Ye, Yang and Zhu, Bin and Cui, Jiaxi and Ning, Munan and Jin, Peng and Yuan, Li},
  booktitle={Proceedings of the 2024 conference on empirical methods in natural language processing},
  pages={5971--5984},
  year={2024}
}

@misc{zhang2024llavanext,
  title={LLaVA-NeXT: A Strong Zero-shot Video Understanding Model},
  url={https://llava-vl.github.io/blog/2024-04-30-llava-next-video/},
  author={Zhang, Yuanhan and Li, Bo and Liu, haotian and Lee, Yong jae and Gui, Liangke and Fu, Di and Feng, Jiashi and Liu, Ziwei and Li, Chunyuan},
  month={April},
  year={2024}
}

@article{wang2024videohallucer,
  title={Videohallucer: Evaluating intrinsic and extrinsic hallucinations in large video-language models},
  author={Wang, Yuxuan and Wang, Yueqian and Zhao, Dongyan and Xie, Cihang and Zheng, Zilong},
  journal={arXiv preprint arXiv:2406.16338},
  year={2024}
}

@article{zhang2024eventhallusion,
  title={Eventhallusion: Diagnosing event hallucinations in video llms},
  author={Zhang, Jiacheng and Jiao, Yang and Chen, Shaoxiang and Zhao, Na and Tan, Zhiyu and Li, Hao and Ma, Xingjun and Chen, Jingjing},
  journal={arXiv preprint arXiv:2409.16597},
  year={2024}
}

@inproceedings{fu2025videomme,
  title={Video-mme: The first-ever comprehensive evaluation benchmark of multi-modal llms in video analysis},
  author={Fu, Chaoyou and Dai, Yuhan and Luo, Yongdong and Li, Lei and Ren, Shuhuai and Zhang, Renrui and Wang, Zihan and Zhou, Chenyu and Shen, Yunhang and Zhang, Mengdan and others},
  booktitle={2025 IEEE/CVF Conference on Computer Vision and Pattern Recognition (CVPR)},
  pages={24108--24118},
  year={2025},
  organization={IEEE}
}

@inproceedings{li2024mvbench,
  title={Mvbench: A comprehensive multi-modal video understanding benchmark},
  author={Li, Kunchang and Wang, Yali and He, Yinan and Li, Yizhuo and Wang, Yi and Liu, Yi and Wang, Zun and Xu, Jilan and Chen, Guo and Luo, Ping and others},
  booktitle={Proceedings of the IEEE/CVF conference on computer vision and pattern recognition},
  pages={22195--22206},
  year={2024}
}

@inproceedings{li2025vidhalluc,
  title={Vidhalluc: Evaluating temporal hallucinations in multimodal large language models for video understanding},
  author={Li, Chaoyu and Im, Eun Woo and Fazli, Pooyan},
  booktitle={2025 IEEE/CVF Conference on Computer Vision and Pattern Recognition (CVPR)},
  pages={13723--13733},
  year={2025},
  organization={IEEE}
}

@inproceedings{kong2025mhbench,
  title={Mhbench: Demystifying motion hallucination in videollms},
  author={Kong, Ming and Zeng, Xianzhou and Chen, Luyuan and Li, Yadong and Yan, Bo and Zhu, Qiang},
  booktitle={Proceedings of the AAAI Conference on Artificial Intelligence},
  volume={39},
  number={4},
  pages={4401--4409},
  year={2025}
}

@article{xiao2026macd,
  title={MACD: Model-Aware Contrastive Decoding via Counterfactual Data},
  author={Xiao, Qixin and Zhou, Kun},
  journal={arXiv preprint arXiv:2602.01740},
  year={2026}
}

@article{huang2025vistadpo,
  title={Vistadpo: Video hierarchical spatial-temporal direct preference optimization for large video models},
  author={Huang, Haojian and Chen, Haodong and Wu, Shengqiong and Luo, Meng and Fu, Jinlan and Du, Xinya and Zhang, Hanwang and Fei, Hao},
  journal={arXiv preprint arXiv:2504.13122},
  year={2025}
}

@article{ding2025pamivdpo,
  title={Pami-vdpo: Mitigating video hallucinations by prompt-aware multi-instance video preference learning},
  author={Ding, Xinpeng and Zhang, Kui and Han, Jianhua and Hong, Lanqing and Xu, Hang and Li, Xiaomeng},
  journal={arXiv preprint arXiv:2504.05810},
  year={2025}
}

@inproceedings{wu2025season,
  title={Season: Mitigating temporal hallucination in video large language models via self-diagnostic contrastive decoding},
  author={Wu, Chang-Hsun and Chang, Kai-Po and Sheng, Yu-Yang and Chung, Hung-Kai and Wang, Kuei-Chun and Wang, Yu-Chiang Frank},
  booktitle={Proceedings of the IEEE/CVF Conference on Computer Vision and Pattern Recognition},
  pages={11096--11105},
  year={2026}
}

@inproceedings{ma2024vistallama,
  title={Vista-llama: Reducing hallucination in video language models via equal distance to visual tokens},
  author={Ma, Fan and Jin, Xiaojie and Wang, Heng and Xian, Yuchen and Feng, Jiashi and Yang, Yi},
  booktitle={2024 IEEE/CVF Conference on Computer Vision and Pattern Recognition (CVPR)},
  pages={13151--13160},
  year={2024},
  organization={IEEE}
}

@inproceedings{chang2026santa,
  title={Mitigating object and action hallucinations in multimodal llms via self-augmented contrastive alignment},
  author={Chang, Kai-Po and Cheng, Wei-Yuan and Huang, Chi-Pin and Yang, Fu-En and Wang, Yu-Chiang Frank},
  booktitle={2026 IEEE/CVF Winter Conference on Applications of Computer Vision (WACV)},
  pages={3172--3181},
  year={2026},
  organization={IEEE}
}

@inproceedings{zhang2026rewatch,
  title={ReWatch-R1: Boosting Complex Video Reasoning in Large Vision-Language Models through Agentic Data Synthesis},
  author={Zhang, Congzhi and Wang, Zhibin and Ma, Yinchao and Peng, Jiawei and Wang, Yihan and Zhou, Qiang and Song, Jun and Zheng, Bo},
  booktitle={International Conference on Learning Representations},
  volume={2026},
  pages={140958--140985},
  year={2026}
}

@inproceedings{liu2024pai,
  title={Paying more attention to image: A training-free method for alleviating hallucination in lvlms},
  author={Liu, Shi and Zheng, Kecheng and Chen, Wei},
  booktitle={European Conference on Computer Vision},
  pages={125--140},
  year={2024},
  organization={Springer}
}

@inproceedings{liu2025vti,
  title={Reducing hallucinations in large vision-language models via latent space steering},
  author={Liu, Sheng and Ye, Haotian and Zou, James Y},
  booktitle={International Conference on Learning Representations},
  volume={2025},
  pages={72402--72419},
  year={2025}
}

@inproceedings{cai2025taae,
 author = {Cai, Jianfeng and Hong, Jiale and Zhang, Zongmeng and Zhou, Wengang and zhannianji and Li, Houqiang},
 booktitle = {Advances in Neural Information Processing Systems},
 doi = {10.52202/085713-2006},
 editor = {D. Belgrave and C. Zhang and H. Lin and R. Pascanu and P. Koniusz and M. Ghassemi and N. Chen},
 pages = {59982--60012},
 publisher = {Curran Associates, Inc.},
 title = {Mitigating Hallucination in VideoLLMs via Temporal-Aware Activation Engineering},
 url = {https://proceedings.neurips.cc/paper_files/paper/2025/file/5683583b88da51c79d2eb263f295286e-Paper-Conference.pdf},
 volume = {38, Main Conference},
 year = {2025}
}

@inproceedings{wang2026ascd,
  title={Ascd: Attention-steerable contrastive decoding for reducing hallucination in mllm},
  author={Wang, Yujun and Bi, Jinhe and Pirk, Soren and Ma, Yunpu and others},
  booktitle={Proceedings of the AAAI Conference on Artificial Intelligence},
  volume={40},
  number={12},
  pages={10306--10314},
  year={2026}
}

@article{xing2026vader,
  title={VADER: Adaptive Debiasing for Hallucination Mitigation in Video Large Language Models},
  author={Xing, Dong and Chen, Jiaxin and Yang, Hang and Liu, Peixun and Yang, Qiushi and Wang, Yuqing},
  journal={arXiv preprint arXiv:2608.08622},
  year={2026}
}

@article{wang2026after,
  title={AFTER: Mitigating the Object Hallucination of LVLM via Adaptive Factual-Guided Activation Editing},
  author={Wang, Tianbo and Ma, Yuqing and Liao, Kewei and Zhang, Zhange and Li, Simin and Guo, Jinyang and Liu, Xianglong},
  journal={arXiv preprint arXiv:2601.01957},
  year={2026}
}

@inproceedings{maaz2024videochatgpt,
  title={Video-chatgpt: Towards detailed video understanding via large vision and language models},
  author={Maaz, Muhammad and Rasheed, Hanoona and Khan, Salman and Khan, Fahad},
  booktitle={Proceedings of the 62nd Annual Meeting of the Association for Computational Linguistics (Volume 1: Long Papers)},
  pages={12585--12602},
  year={2024}
}

@inproceedings{zhou2026cctvbench,
  title={CCTVBench: Contrastive Consistency Traffic VideoQA Benchmark for Multimodal LLMs},
  author={Zhou, Xingcheng and Guo, Hao and Song, Rui and Zimmer, Walter and Liu, Mingyu and Schamschurko, Andr{\'e} and Cao, Hu and Knoll, Alois},
  booktitle={Findings of the Association for Computational Linguistics: ACL 2026},
  pages={21665--21684},
  year={2026}
}

@article{wu2026omniworldbench,
  title={Omni-worldbench: Towards a comprehensive interaction-centric evaluation for world models},
  author={Wu, Meiqi and Cai, Zhixin and Zhao, Fufangchen and Feng, Xiaokun and Dang, Rujing and Song, Bingze and Tian, Ruitian and Zhu, Jiashu and Lei, Jiachen and Dou, Hao and others},
  journal={arXiv preprint arXiv:2603.22212},
  year={2026}
}

@inproceedings{huang2024opera,
  title={Opera: Alleviating hallucination in multi-modal large language models via over-trust penalty and retrospection-allocation},
  author={Huang, Qidong and Dong, Xiaoyi and Zhang, Pan and Wang, Bin and He, Conghui and Wang, Jiaqi and Lin, Dahua and Zhang, Weiming and Yu, Nenghai},
  booktitle={2024 IEEE/CVF Conference on Computer Vision and Pattern Recognition (CVPR)},
  pages={13418--13427},
  year={2024},
  organization={IEEE}
}

@inproceedings{suo2025octopus,
  title={Octopus: Alleviating hallucination via dynamic contrastive decoding},
  author={Suo, Wei and Zhang, Lijun and Sun, Mengyang and Wu, Lin Yuanbo and Wang, Peng and Zhang, Yanning},
  booktitle={2025 IEEE/CVF Conference on Computer Vision and Pattern Recognition (CVPR)},
  pages={29904--29914},
  year={2025},
  organization={IEEE}
}

@inproceedings{huo2025sid,
  title={Self-introspective decoding: Alleviating hallucinations for large vision-language models},
  author={Huo, Fushuo and Xu, Wenchao and Zhang, Zhong and Wang, Haozhao and Chen, Zhicheng and Zhao, Peilin},
  booktitle={International Conference on Learning Representations},
  volume={2025},
  pages={24272--24295},
  year={2025}
}

@article{zhang2025adaqr,
  title={Your dense retriever is secretly an expeditious reasoner},
  author={Zhang, Yichi and Bai, Jun and Cai, Zhixin and Qin, Shuhan and Chen, Zhuofan and Guan, Jinghua and Rong, Wenge},
  journal={arXiv preprint arXiv:2510.21727},
  year={2025}
}

@article{cai2026xetrieval,
  title={Xetrieval: Mechanistically Explaining Dense Retrieval},
  author={Cai, Zhixin and Bai, Jun and Liu, Yang and Li, Jiaqi and Zhang, Yichi and Li, Taichuan and Chen, Zhuofan and Jia, Zixia and Zheng, Zilong and Rong, Wenge},
  journal={arXiv preprint arXiv:2605.29507},
  year={2026}
}

@inproceedings{kang2025visualsink,
 author = {Kang, Seil and Kim, Jinyeong and Kim, Junhyeok and Hwang, Seong Jae},
 booktitle = {International Conference on Learning Representations},
 editor = {Y. Yue and A. Garg and N. Peng and F. Sha and R. Yu},
 pages = {87676--87703},
 title = {See What You Are Told: Visual Attention Sink in Large Multimodal Models},
 url = {https://proceedings.iclr.cc/paper_files/paper/2025/file/da8a39bc39ae1c89dd6ebb1e3bcbb3f3-Paper-Conference.pdf},
 volume = {2025},
 year = {2025}
}
\clearpage
\appendix
\section{Evaluation Details}
\label{app:benchmarks}

\subsection{VideoHallucer}

VideoHallucer evaluates intrinsic and extrinsic hallucinations across five categories: object--relation, temporal, semantic detail, extrinsic factual, and extrinsic non-factual hallucinations \citep{wang2024videohallucer}.
Our evaluation covers 948 videos and 976 paired tests, requiring 1,952 yes/no responses.
Each test pairs a basic question with a hallucination question and is counted as correct only when both answers are correct.
We report the macro-average of paired accuracy across the five categories.

\subsection{EventHallusion}

EventHallusion probes event hallucinations through misleading questions and uncommon events that challenge language priors and learned scene--event associations \citep{zhang2024eventhallusion}.
The benchmark contains 397 videos in three groups: 109 \emph{Entire} videos depicting rare events throughout, 193 \emph{Mix} videos interleaving common and rare events, and 95 \emph{Misleading} videos with questions suggesting events absent from the video.
We evaluate all 409 binary selection questions across the three groups and 302 free-form descriptions for Entire and Mix.
Selection responses are scored using the benchmark's official answer parser.
All descriptions are evaluated using \texttt{gpt-5.6-luna} with the official judging prompt, which checks agreement with the annotated event.
The overall score is the count-weighted aggregate of selection and description scores.

\subsection{Video-MME}

Video-MME comprises 900 videos and 2,700 human-annotated multiple-choice questions spanning six content domains, 30 subcategories, and 12 question types \citep{fu2025videomme}.
Each video has three questions, and each duration group contains 300 videos: short (under 2 minutes), medium (4--15 minutes), and long (30--60 minutes).
We evaluate all three groups using video frames without audio or subtitles.

\paragraph{Scoring and answer coverage.}
\label{app:coverage}

Following the official Video-MME evaluator,\footnote{\url{https://github.com/thanku-all/parse_answer}} we compute accuracy as correct/answered, where answered counts responses with a valid extracted option.
Responses without an extracted option are excluded from the accuracy denominator.

\begin{table}[t]
\centering
\small
\begin{tabular}{lrr}
\toprule
Method & Correct/answered & Coverage (\%) \\
\midrule
\multicolumn{3}{l}{\textbf{Qwen2.5-VL-3B}} \\
Vanilla & 1400/2700 & 100.00 \\
DINO-HEAL & 1401/2700 & 100.00 \\
TCD & 1404/2700 & 100.00 \\
MotionCD & 1227/2700 & 100.00 \\
MACD & 1365/2700 & 100.00 \\
\method{} & 1417/2700 & 100.00 \\
\midrule
\multicolumn{3}{l}{\textbf{Qwen2.5-VL-7B}} \\
Vanilla & 1445/2700 & 100.00 \\
DINO-HEAL & 1455/2700 & 100.00 \\
TCD & 1456/2700 & 100.00 \\
MotionCD & 1255/2700 & 100.00 \\
MACD & 1413/2699 & 99.96 \\
\method{} & 1474/2700 & 100.00 \\
\midrule
\multicolumn{3}{l}{\textbf{Video-LLaVA-7B}} \\
Vanilla & 707/2700 & 100.00 \\
DINO-HEAL & 737/2700 & 100.00 \\
TCD & 717/2700 & 100.00 \\
MotionCD & 781/2699 & 99.96 \\
MACD & 773/2699 & 99.96 \\
\method{} & 888/2698 & 99.93 \\
\midrule
\multicolumn{3}{l}{\textbf{LLaVA-NeXT-Video-7B}} \\
Vanilla & 806/2697 & 99.89 \\
DINO-HEAL & 810/2700 & 100.00 \\
TCD & 836/2700 & 100.00 \\
MotionCD & 868/2699 & 99.96 \\
MACD & 852/2692 & 99.70 \\
\method{} & 999/2697 & 99.89 \\
\bottomrule
\end{tabular}
\caption{Video-MME answer coverage for the main comparison, out of 2,700 questions per run. Accuracy follows the official evaluator's correct/answered convention.}
\label{tab:coverage}
\end{table}

Table~\ref{tab:coverage} reports answer coverage, defined as the fraction of questions with a valid extracted option.
Answer coverage is nearly complete across models and methods, so excluding unparseable responses has a negligible effect on the main comparison.

\subsection{MVBench}

MVBench evaluates temporal video understanding through 20 multiple-choice tasks, including action sequencing, object interaction, motion perception, episodic reasoning, and counterfactual inference \citep{li2024mvbench}.
The full evaluation set contains 4,000 questions, with 200 questions per task.
Answers are scored against the ground-truth option, with unparseable responses counted as incorrect.
We report overall question accuracy, equivalent to the mean task accuracy because all tasks have the same size.

\section{Ablation Details}
\label{app:implementation}

\paragraph{Magnitude-matched perturbations.}
For example $i$, let $r_i=\RMS(H_i^--H_i)/\RMS(H_i)$ be \method{}'s realized relative displacement and $s_i=\RMS(H_i)$.
\textit{IsoPert} samples an isotropic Gaussian direction $G_i$ and constructs $H_i^-=H_i+s_i r_i G_i/\RMS(G_i)$.
\textit{RadCon} instead uses $H_i^-=(1-r_i)H_i$.
Both match the perturbation magnitude of \method{}, isolating the effect of its optimized direction.

\paragraph{Query reconstruction.}
\textit{QueryRec} applies MACD's question-reconstruction objective to the visual representations \citep{xiao2026macd}:
\begin{equation}
 J_{\mathrm{rec}}(H';x)=-\frac{1}{|\mathcal I|}\sum_{j\in\mathcal I}\log p_\theta(x_j\mid H',x_{<j}),
\end{equation}
where $\mathcal I$ indexes the user-instruction tokens, including answer options when present, but excluding chat-template and video tokens.
For EventHallusion descriptions, the instruction is ``Please describe this video in detail.''
QueryRec uses the same perturbation constraint and projected update rule as \method{}, selecting the iterate with the highest reconstruction loss.

\paragraph{Attention calibration.}
\textit{AttnCal} subtracts a common nonnegative bias $\kappa$ from visual-key attention logits at layer $\ell$ in the auxiliary branch.
For a head with original visual attention mass $m_h$, the adjusted mass is
\begin{equation}
 m_h(\kappa)=\frac{e^{-\kappa}m_h}{1-m_h+e^{-\kappa}m_h}.
\end{equation}
We choose $\kappa$ so that $\frac{1}{n_h}\sum_hm_h(\kappa)=m_\ell(H^-,x)$, matching \method{}'s attention reduction without altering the input visual representations.
The same bias is applied at the last prompt position and subsequent answer positions throughout generation.

\section{Mechanistic Analysis Details}
\label{app:diagnostics}

Attention retention in Table~\ref{tab:contrastive_effects} is measured as $m_\ell(H^-,x)/m_\ell(H,x)$.
At the first answer position, each answer label is represented by its highest-scoring single-token form under the original branch.
The margins defined in Eq.~\ref{eq:branch_margins} use raw logits before repetition penalty.
All methods share the candidate pairs and evaluation subset selected using the original branch's plausibility mask.

\section{Effect of Question Conditioning}
\label{app:question_transfer}

To assess the role of question-specific optimization, we compare \method{} with perturbations optimized for another question about the same video.
The comparison uses all 2,700 Video-MME questions for Qwen2.5-VL-3B and Video-LLaVA-7B.

\begin{table}[!htbp]
\centering
\small
\setlength{\tabcolsep}{3pt}
\begin{tabular}{lrrrr}
\toprule
Model & Vanilla & Transfer & \method{} & $\Delta$ \\
\midrule
Qwen2.5-VL-3B & 51.85 & 52.30 & \textbf{52.48} & +0.19 \\
Video-LLaVA-7B & 26.19 & 30.02 & \textbf{32.91} & +2.89 \\
\bottomrule
\end{tabular}
\caption{Effect of question conditioning on Video-MME (\%).
Transfer uses perturbations optimized for another question about the same video.
$\Delta$ is \method{} minus Transfer (pp).}
\label{tab:question_transfer}
\end{table}

\paragraph{Perturbation transfer.}
For each video, we cyclically assign each of its three questions to another as the optimization source, including the source question's answer options.
We optimize with the same attention objective and update rule, then rescale the resulting perturbation to match \method{}'s realized RMS for the target question.
The transferred perturbation is used as the negative input when answering the target question with its original prompt.
This comparison isolates question conditioning from video content and perturbation magnitude.

\paragraph{Results.}
The source and target questions concern the same video and can share semantic content and visual evidence, which helps explain why transferred perturbations remain beneficial over Vanilla.
As shown in Table~\ref{tab:question_transfer}, optimizing for the target question yields further gains on both models, supporting the value of question-specific negative construction.

\end{document}